\documentclass[letterpaper, 10 pt, conference]{ieeeconf}

\IEEEoverridecommandlockouts                              % This command is only needed if 
\usepackage{icra}

\title{\LARGE \bf Sight Over Site: Perception-Aware Reinforcement Learning\\ for Efficient Robotic Inspection}

\usepackage{comment}

\author{Richard Kuhlmann$^{1, *}$, Jakob Wolfram$^{1, *}$, Boyang Sun$^2$, Jiaxu Xing$^3$, \\Davide Scaramuzza$^3$, Marc Pollefeys$^{2, 4}$, Cesar Cadena$^1$% <-this % stops a space
\thanks{$^{*}$ These authors contributed equally to this work.}% <-this % stops a space
\thanks{$^1$ Robotics Systems Lab, ETH Zurich, $^2$ Computer Vision and Geometry Group, ETH Zurich $^3$ Robotics and Perception Group, University of Zurich, 
$^4$ Microsoft Mixed Reality and AI Lab}
\thanks{Contact: \tt\small rkuhlmann@ethz.ch}
\thanks{This work was supported by Huawei Tech R\&D (UK) through a research funding agreement, by the Swiss National Science Foundation (SNSF) as part of project No.227617, and by the European Union’s Horizon Europe Research and Innovation Programme under grant agreement No. 101120732 (AUTOASSESS)}
}
\begin{document}

\maketitle

\begin{abstract}
    Autonomous inspection is a central problem in robotics, with applications ranging from industrial monitoring to search-and-rescue. 
    Traditionally, inspection has often been reduced to navigation tasks, where the objective is to reach a predefined location while avoiding obstacles.
    However, this formulation captures only part of the real inspection problem.
    In real-world environments, the inspection targets may become visible well before their exact coordinates are reached, making further movement both redundant and inefficient.
    What matters more for inspection is not simply arriving at the target’s position, but positioning the robot at a viewpoint from which the target becomes observable.
    In this work, we revisit inspection from a perception-aware perspective.
    We propose an end-to-end reinforcement learning framework that explicitly incorporates target visibility as the primary objective, enabling the robot to find the shortest trajectory that guarantees visual contact with the target without relying on a map. 
    The learned policy leverages both perceptual and proprioceptive sensing and is trained entirely in simulation, before being deployed to a real-world robot. 
    We further develop an algorithm to compute ground-truth shortest inspection paths, which provides a reference for evaluation. 
    Through extensive experiments, we show that our method outperforms existing classical and learning-based navigation approaches, yielding more efficient inspection trajectories in both simulated and real-world settings.

    The project is avialable at \href{sight-over-site.github.io}{https://sight-over-site.github.io/}
\end{abstract}
\section{Introduction}
\label{sec:introduction}
Many real-world robotic applications fundamentally require a robot to obtain visual access to a target.
Examples include inspecting devices~\cite{xing2023autonomous, ollero2024application}, reading measurement displays, detecting visual signals in industrial settings~\cite{lattanzi2017review}, tracking objects and people for service robots, enabling human-robot collaboration~\cite{chen20243d,Munaro2014FastRP}, and supporting search-and-rescue operations~\cite{der2021roboa,schwaiger2024semi}. 
The core objective in these scenarios is similar and can be collectively described as autonomous inspection~\cite{inspect-planning-bircher, belief-behavior-graph, semantics-aware-exploration}.

% Unlike autonomous navigation, which focuses on reaching a specific location, inspection prioritizes acquiring visual information about specific targets. The goal is to observe these targets as efficiently as possible, rather than merely moving closer to them~\cite{inspect-planning-bircher, belief-behavior-graph, semantics-aware-exploration}.
%
\begin{figure}[t]
    \centering    
    \vspace{0.5em}
    \includegraphics[width=0.98\linewidth]{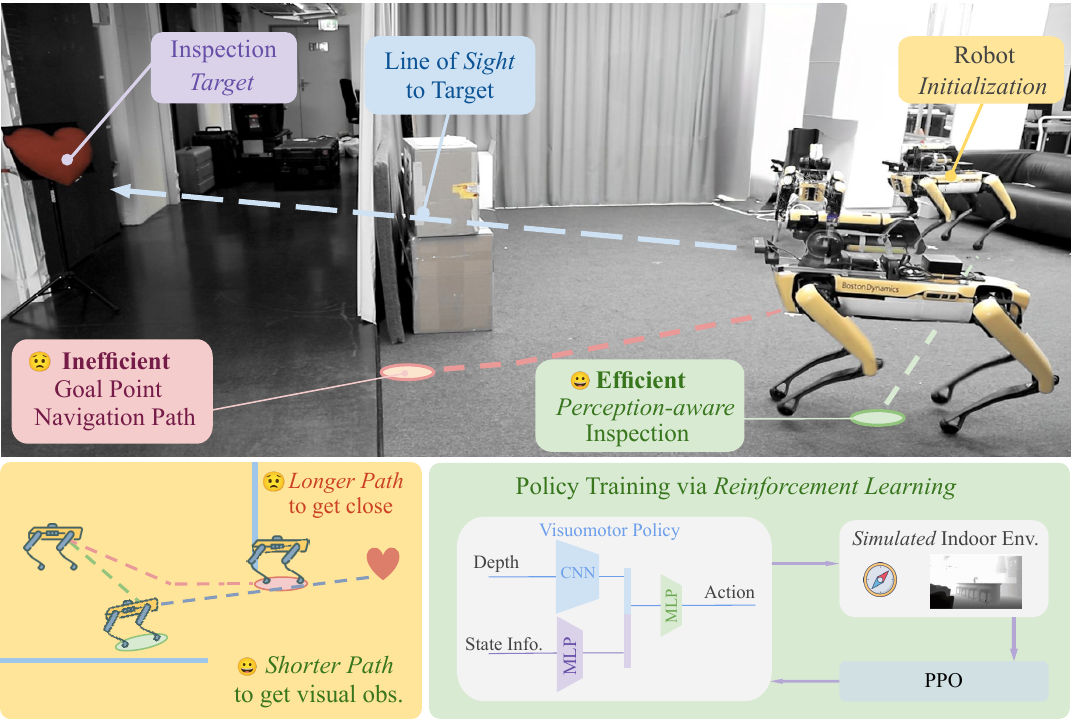}
    \caption{To obtain visual access to a given target, the \textit{inspection} policy (green path) results in a shorter trajectory compared to the \textit{navigation} policy (red path). Our proposed RL-based policy (bottom right) takes egocentric depth input along with the target and robot state to achieve efficient inspection.
    }
    \label{fig:init_figure}
\end{figure}
Despite its importance, inspection has received limited attention as a standalone objective. Most existing approaches treat inherently inspection-driven tasks as navigation problems, where the focus is on reaching a specific location~\cite{Wijmans2020DD-PPO, shah2022gnm, batra2020objectnav, yang2022far, iplanner}, rather than explicitly optimizing for target visibility. In such cases, even though the goal is to observe the target, the robot primarily focuses on moving closer to it. Navigation-based approaches can occasionally succeed in inspection tasks, as proximity to the target increases the likelihood of observing it. However, this strategy often leads to a sub-optimal behavior. As shown in~\ref{fig:init_figure}, the most direct path to achieve visibility can be significantly shorter than the trajectory chosen by a well-trained navigation policy. This discrepancy arises because navigation approaches typically do not consider visibility in their decision-making process. In real-world scenarios, this limitation can cause significant inefficiencies, such as the need to sequentially visit multiple viewpoints in cluttered environments.

% One way to formulate target-inspection-related tasks is as navigation problems, where the robot is encouraged to reach visibility of the target by moving towards it. While navigation-based approaches can succeed in inspection tasks -- since approaching the target often increases the chance of observing it -- this strategy may lead to sub-optimal solutions.
% As illustrated in figure~\ref{fig:init_figure}, the most direct path to achieve visibility may be significantly shorter than the one chosen by a well-trained navigation policy. This discrepancy arises because navigation approaches do not prioritize visibility. In real-world scenarios, this limitation can lead to inefficient inspection paths.

Inspired by these observations, we propose a perception-aware reinforcement learning (RL) framework for efficient robotic inspection. Our approach explicitly incorporates the visibility objective into the RL training process, encouraging the robot to obtain visibility of a target 3D location without relying on a map. The policy takes as input depth sensor readings, the relative distance and angle to the target, and the agent’s last three actions, and outputs successive points of interest relative to the robot’s position. A low-level controller then executes the corresponding joint commands to reach these points. We train the policy entirely in simulation, and it generalizes well to both unseen simulated environments and a real-world quadrupedal robot (Boston Dynamics Spot). Our policy outperforms current state-of-the-art point-goal navigation methods and demonstrates strong performance in collision avoidance. Notably, the learned policy exhibits an emergent ability to balance navigation and inspection. When the target is far away, it prioritizes navigating towards the vicinity of the goal. Once nearby, it shifts focus towards reaching a feasible viewpoint as efficiently as possible.
%
% \subsection*{Contributions}
In conclusion, the main contributions of this work are:
\begin{itemize}
    \item A comprehensive definition and formulation of the perceptive inspection task, along with a modeling framework for evaluating the inspection process.
    \item A map-free, RL-based policy that leverages egocentric depth input and the relative pose to the target for efficient autonomous inspection. 
    \item Quantitative evaluation of the proposed system against a state-of-the-art RL-based navigation approach, conducted in both simulation and real-world settings.
\end{itemize}

\section{Related Work}
\label{sec:relatedworks}
Perception-aware navigation problems in unknown environments have been studied extensively in the past~\cite{visualnav-Bonin,objectnav_revisited,liu2025aligning,zhang2022survey}.
This general class of problems includes a range of tasks, such as exploration in unknown spaces~\cite{dai2020fast,cao2021tare,boysun2025frontiernet}, navigating to a specified location~\cite{Wijmans2020DD-PPO,iplanner,yang2022far}, or reaching specific objects or semantic regions~\cite{gervet2023navigating,tao2022seer}. 
To address these different tasks, a variety of methods have been developed. 
A major line of work focuses on gradually constructing and updating a global map, which is then used to guide planning -- either by selecting informative regions such as frontiers~\cite{frontier-exp,ramakrishnan2022poni} or by predicting next-best views to observe a target directly~\cite{Schmid20ActivePlanning,semantics-aware-exploration}. 
More recently, with the rapid advancement of large-scale multi-modality models, several approaches have explored integrating high-level reasoning from these models into the mapping and decision-making process~\cite{chen2023not,yokoyama2024vlfm,shah2023vint,shah2025foresightnav}. 
However, most existing methods focus on reaching a target location, often without explicitly considering whether the target becomes visible to the robot. 
Incorporating visibility into the objective, either through hand-designed information gain measures or queries to pretrained language models, remains an open research question. Furthermore, to achieve robustness, many of these approaches rely on maintaining an explicit global map or long-term memory.

In parallel, another line of work leverages reinforcement learning to solve navigation tasks, avoiding modular system designs and taking advantage of large-scale, high-fidelity simulators.
In particular, the poing-goal navigation task has seen rapid progress using RL-based approaches. 
Policies trained on large-scale simulation data have demonstrated near-perfect performance~\cite{Wijmans2020DD-PPO}. 
Follow-up work has further improved performance in more challenging settings by developing better training schemes and network architectures~\cite{iplanner}, incorporating semantics~\cite{roth2024viplanner,object-goal-navigation}, extending to object-goal navigation~\cite{xie2025naviformer}, and enabling multi-robot coordination~\cite{morad2025language,feng2025safe}. 
Recent RL-based approaches have also improved generalization across tasks and environments, and reduced the sim-to-real gap~\cite{cai2025navdp,boiteau2025model}. Despite this progress, few studies have investigated visibility as a core component of navigation. 

Building on this body of work, we study the problem of \emph{perception-aware inspection}. 
Specifically, we consider the task of navigating in an unknown, map-less environment to reach a vantage point from which the target is visible, using only egocentric depth input.
In contrast to common navigation approaches, our formulation does not require the robot to reach the physical location of the target. 
Instead, the emphasis is placed on perception—finding a feasible viewpoint from which the target can be observed. 
This perspective provides a new lens for understanding and addressing inspection tasks in realistic, map-free settings.
\section{Methodology}
\label{sec:method}
% In this section, we define the perceptive inspection task in Sec.~\ref{subsec:task_def}, introduce the shortest inspection path as a ground-truth reference in Sec.~\ref{subsec:ground_truth}, and describe the environment setup in Sec.~\ref{subsec:env_setup}. 
% %
% We then present the agent’s observation space in Sec.~\ref{subsec:obs}, action space in Sec.~\ref{subsec:act}, reward design in Sec.~\ref{subsec:reward}, and policy architecture in Sec.~\ref{subsec:architecture} used for training and evaluation.
%
\subsection{Task Definition}
\label{subsec:task_def}
The objective in our \emph{perceptive inspection} task is for a robotic agent to autonomously discover the shortest path to a vantage point from which a designated target becomes observable.
A target is considered observable if it lies within the agent’s field of view and the line of sight is free of occlusions caused by obstacles.
In the classical point-goal navigation setting, the goal location is explicitly provided.
In inspection, by contrast, the vantage point from which the target becomes visible is not known beforehand.
The agent must therefore determine, from its sensory inputs, how to move to reveal the target.
This makes the problem challenging, as it requires a tight integration of perception, planning, and control to actively search for informative viewpoints while avoiding collisions.
In addition, motivated by prior work demonstrating that agents can operate effectively without relying on memory-intensive mapping~\cite{Wijmans2020DD-PPO}, we frame the problem using end-to-end reinforcement learning. 
This choice avoids the overhead of explicit map construction, enabling agents to scale to large environments while still ensuring collision-free trajectories.
\subsection{Formulating Evaluation Framework for Inspection}
\label{subsec:ground_truth}
To evaluate the quality and efficiency of a robot’s inspection trajectory, it is 
necessary to establish a ground-truth reference, namely the \emph{shortest inspection path}, as illustrated in Alg.~\ref{alg:shortest_path}.
We therefore propose an approach to determine this optimal path 
under full map observability in our perceptive inspection setting.

We define $l$ as the shortest collision-free path from the robot’s start position 
$\mathbf{S} \in \mathbb{R}^3$ to an inspection goal 
$\mathbf{G}_\text{opt} \in \mathbb{R}^3$ in the domain $D \subset \mathbb{R}^3$, 
from which the target $\mathbf{T} \in \mathbb{R}^3$ is observable.
Given privileged access to a 2D top-down occupancy grid of the environment, the 3D domain $D$ can be projected onto a planar subset  $D_\text{proj} \subset \mathbb{R}^2$. 
The projected domain is partitioned into traversable space $D_\text{free} \subset D_\text{proj}$ and occupied space 
$D_\text{occ} = D_\text{proj} \setminus D_\text{free}$. 
Candidate inspection goal positions are grid cells $\mathbf{C} \in D_\text{free}$ that satisfy the following constraints:
(i) \emph{Traversability:} $\mathbf{C} \in D_\text{free}$.
(ii) \emph{Sensor range:} 
    $\|\mathbf{T}_\text{proj} - \mathbf{C}\|_2 \leq \delta$, where 
    $\mathbf{T}_\text{proj}$ is the 2D projection of the target.
(iii) \emph{Visibility:} The line of sight between $\mathbf{C}$ and $\mathbf{T}$ is unobstructed.

To determine visibility in 3D, we simulate the robot’s depth sensor at grid cell $\mathbf{C}$.
%with pose $(\mathbf{R}, \mathbf{t}) \in SE(3)$. 
%
Let $(u,v)$ be the pixel coordinates of the target’s projection on the camera plane. 
Here we denote the measured depth at $(u,v)$ by $d_\text{depth}(\mathbf{u})$ and the 
Euclidean distance from $\mathbf{C}$ to the target’s projection by 
\begin{equation}
d_\text{target} = \|\mathbf{T}_\text{proj} - \mathbf{C}\|_2.
\end{equation} 
The target is observable if and only if
$d_\text{depth}(\mathbf{u}) \geq d_\text{target} > 0.$
Otherwise, the line of sight is occluded by an obstacle.

Since the robot’s translational motion is constrained to the $xy$-plane, the shortest inspection path can be computed on the 2D occupancy grid. 
Let the set of candidate inspection goal cells be:
\[
    G = \{ \mathbf{G} \in D_\text{free} \;\; | \;\; 
    \mathbf{G} \text{ satisfies constraints (i)--(iii)} \}.
\]
We then apply the A* search algorithm~\cite{A*} to find the shortest path from 
$\mathbf{S}$ to any $\mathbf{G} \in G$. The heuristic function is
\begin{equation}
    h(\mathbf{C}) = \min_{\mathbf{G} \in G} \|\mathbf{C} - \mathbf{G}\|_2,
\end{equation}
i.e., the Euclidean distance from the current cell $\mathbf{C}$ to the nearest candidate goal.
When A* expands a cell in $G$, the corresponding shortest 
inspection path $l$ is obtained. 
Thus, a single execution of A* is sufficient to 
determine the optimal path for a given start-target pair.
\begin{algorithm}[t]
\caption{Shortest Perceptive Inspection Path}
\label{alg:shortest_path}
\KwIn{Occupancy grid $D_\text{proj}$, start $\mathbf{S}$, target $\mathbf{T}$, sensor range $\delta$}
\KwOut{Shortest inspection path $l$}

Project target: $\mathbf{T}_\text{proj} \gets \text{proj}(\mathbf{T})$ \;
Define free space $D_\text{free}$ from $D_\text{proj}$ \;

$G \gets \emptyset$ \tcp*{Candidate goal set}
\For{$\mathbf{C} \in D_\text{free}$}{
    \If{$\|\mathbf{T}_\text{proj} - \mathbf{C}\|_2 \leq \delta$ \textbf{and} visible($\mathbf{C}, \mathbf{T}$)}{
        $G \gets G \cup \{\mathbf{C}\}$ \;
    }
}

Run A* from $\mathbf{S}$ to goal set $G$,
using heuristic $h(\mathbf{C}) = \min_{\mathbf{G} \in G} \|\mathbf{C} - \mathbf{G}\|_2$\\
$l \gets$ resulting shortest path \;
\Return $l$
\end{algorithm}

However, following this approach to use the ground-truth path as a reward signal in an RL training pipeline is impractical, since the set of candidate inspection goal points $G$ changes every time the agent moves. This requires repeatedly running A* search on the occupancy grid along with visibility checks for the candidate goal cells.
Therefore, in this work, we primarily employ the shortest perceptive inspection path for benchmarking. 
For RL policy training, instead of inefficiently recomputing ground-truth paths, we incorporate the same underlying principle into the environment and reward design to ensure efficient policy optimization, as detailed in the following subsections.

\subsection{Environment Setup}
\label{subsec:env_setup}
From the environment setup, we follow several design choices established in prior PointGoal Navigation works~\cite{objectnav_revisited, mapping_necessary, Wijmans2020DD-PPO}.
At the beginning of each episode, the agent is initialized at a random position in the domain $D$ with a random orientation.
At each time step $k$, the agent observes its relative position $\Delta \mathbf{p}_k$ and orientation $\Delta \theta_k$ with respect to the target, together with an ego-centric depth image $\bm{I}_k \in \mathbb{R}^{H \times W}$.
Based on the state and perception, the agent selects its next action.
\begin{figure}[t]
    \centering
    \includegraphics[width=\linewidth]{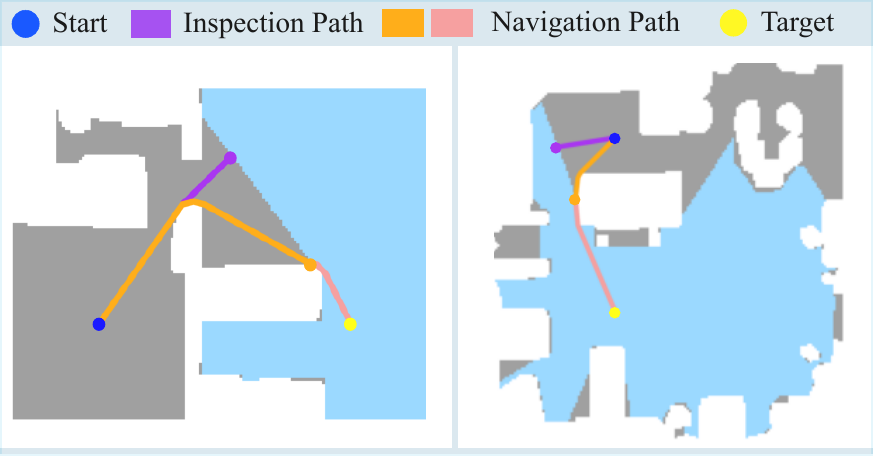}
    \vspace{-0.4cm}
    \caption{
   Examples of shortest inspection path (purple) and the shortest navigation path, shown in orange until reaching a cell with target visibility and in pink thereafter. Traversable areas are depicted in gray, obstacles in white, and cells with target visibility in light blue.
    }
    \label{fig:qualitative_results_gt}
    \vspace{-0.4cm}
\end{figure}
\subsection{Observation Space}
\label{subsec:obs}
At each time step $k$, the agent receives an observation $\mathbf{o}_k$.
Formally, the observation $\mathbf{o}_k$ consists of four components:
(i) the vector from the agent to the target $\Delta \mathbf{p}_k \in \mathbb{R}^2$,  
(ii) the relative orientation to the target $\Delta \theta_k \in [-\pi, \pi)$,  
(iii) an ego-centric depth image $\mathbf{I}_k \in \mathbb{R}^{H \times W}$,  
(iv) and the past three actions chosen by the agent, $\mathbf{u}_{k-1}, \mathbf{u}_{k-2}, \mathbf{u}_{k-3}$.  

The position vector from the agent to the target $\Delta \mathbf{p}_k$ provides the agent with geometric context and serves as a high-level navigation cue. 
Importantly, unlike in PointGoal navigation, this information does not reveal the relative position to the true target of the inspection task, which is to identify the shortest path to a point in the domain $D$ from which the target becomes observable, but the relative position to the target to inspect.
The relative orientation $\Delta \theta_k$ is defined as the angular difference between the agent’s heading direction and the direction of the vector $\Delta \mathbf{p}_k$.
Since the agent is restricted to rotations around its vertical axis, this definition is sufficient and does not require considering the full $SO(3)$ rotation group.
The ego-centric depth image $\mathbf{I}_k$ provides a local perceptual view of the environment from the agent’s perspective.
Depth is used instead of RGB to emphasize geometric structure, which is particularly important in indoor environments characterized by narrow hallways and frequent occlusions. 
This pre-processing preserves the most relevant spatial features while maintaining tractability for policy learning.
To incorporate temporal context and thereby enhance decision-making, the policy input is augmented with the three most recent actions executed by the agent.

\begin{figure*}[t]
    \centering
    % \includesvg[width=0.98\linewidth]{figures/figureedit_paper.svg}
    \includegraphics[width=0.98\linewidth]{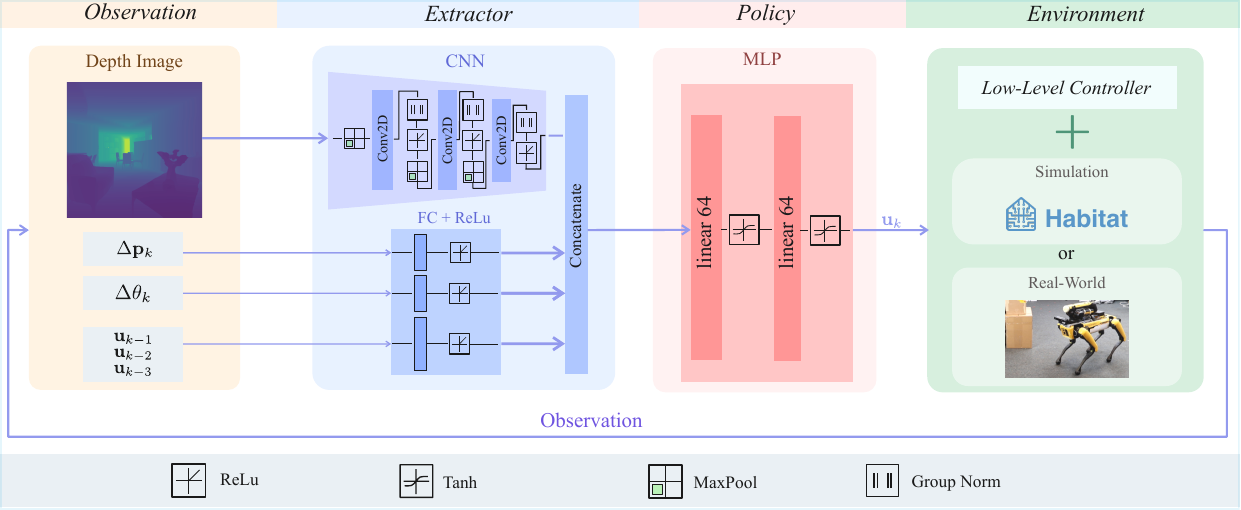}
    \caption{Architecture of the pipeline for training and evaluation. A simulated environment or the robot's sensors provide the observations, which consist of the current ego-centric depth image $\mathbf{I}_k$, the position vector from the agent to the target $\Delta \mathbf{p}_k$, the relative orientation of the agent to the target $\Delta \theta_k$ and the past three actions the agent chose $\mathbf{u}_{k-1}, \mathbf{u}_{k-2}, \mathbf{u}_{k-3}$. Features are extracted from the depth image using a three-layer Convolutional Neural Network (CNN), while the remaining inputs are processed using Single-Layer Perceptrons (SLPs). The next action is selected from a continuous action space by a MLP policy and further executed by a low-level-controller. The policy is trained using Proximal Policy Optimization (PPO).}
    \label{fig:architecture}
    \vspace{-0.5cm}
\end{figure*}

\subsection{Action Space}
\label{subsec:act}
Given the sensor input at time step $k$, the agent learns to predict the next point of interest relative to its current pose in body frame.
The agent's continuous action space $U$ at time step $k$ is represented by the vector $\mathbf{u}_k = [u^{(1)}_k, u^{(2)}_k, u^{(3)}_k]^\top$.
Action $u^{(1)}_k \in [0, 1]$ serves as an activation function and allows the agent to choose between moving forward and rotating in place:
$$
u^{(1)}_k =
\begin{cases}
    \text{move forward} & \text{if } u^{(1)}_k \leq 0.5. \\
    \text{rotate in place} & \text{if } u^{(1)}_k > 0.5.
\end{cases}
$$
By using a continuous range for the activation value $u^{(1)}_k$ instead of a binary activation variable, we get insights into the uncertainty of the policy.
If the policy is uncertain about whether to move forward or rotate in place at time step $k$, the value of $u^{(1)}_k$ remains close to $0.5$.

The components $u^{(2)}_k \in [0, 1]$ and $u^{(3)}_k \in [-1, 1]$ specify the normalized magnitudes for forward translation and rotation, respectively.
These values are subsequently scaled by their respective maximum limits, $s_\text{max} = 0.35\,\text{m}$ for translation and $\phi_\text{max} = \frac{\pi}{4}\,\text{rad}$ for rotation, to obtain the next point of interest.
The decoupling of translation and rotation leads to more robust training and subsequently better policies.

In simulation, the agent's motion is governed by a linearized dynamics model.
The agent's state at time step $k$ is given by its position $\mathbf{p}_k = [x_k, y_k]^\top$ and orientation $\theta_k$ in the world frame. 
The update rule is defined as
\begin{align*}
    \mathbf{p}_{k+1} &= 
    \begin{cases}
        \mathbf{p}_k + s_k \begin{bmatrix} \cos(\theta_k) \\ \sin(\theta_k) \end{bmatrix} & \text{if } u^{(1)}_k \leq 0.5. \\
        \mathbf{p}_k & \text{if } u^{(1)}_k > 0.5.
    \end{cases}, \\[2ex]
    \theta_{k+1} &=
    \begin{cases}
        \theta_k & \text{if } u^{(1)}_k \leq 0.5. \\
        \theta_k + \phi_k & \text{if } u^{(1)}_k > 0.5.
    \end{cases},
\end{align*}
where $s_k = u^{(2)}_k \cdot s_\text{max}$ is the scaled forward translation and $\phi_k = u^{(3)}_k \cdot \phi_\text{max}$ is the scaled rotation.
Thus, at each time step $k$, the agent either moves forward in the direction of its current orientation or rotates in place, depending on the value of $u^{(1)}_k$.

In real-world deployment, a low-level controller is used to convert the next point of interest in the body frame into joint commands to be executed by the robot.
Since our robot’s motion during both training and deployment is not designed for high-speed operation, neglecting the implementation of a low-level controller does not affect real-world performance and even reduces training time.

\subsection{Reward}
\label{subsec:reward}
As the agent seeks to maximize the discounted cumulative reward, the design of the reward function is crucial for effective policy learning.
Table~\ref{tab:rewards} provides an overview of the reward terms.
\renewcommand{\arraystretch}{1.3}
\begin{table}[b]
\caption{Overview of Reward, categorized into sparse and dense types.}
\centering
\begin{tabular}{lll}
\toprule
\textbf{Name} & \textbf{Value} & \textbf{Type} \\
\midrule
\grayrow
$r_{k,T}$ & 
$\begin{cases}
10 & \text{if successful termination} \\
-10 & \text{else}
\end{cases}$ & Sparse \\
$r_{k,\text{orient}}$ & 
$\begin{cases}
0.2 \cdot \Delta\tilde{\mathbf{p}}_k \cdot \mathbf{n}_k & \text{if } \Delta\tilde{\mathbf{p}}_k \cdot \mathbf{n}_k > 0 \\
\Delta\tilde{\mathbf{p}}_k \cdot \mathbf{n}_k & \text{else}
\end{cases}$ & Dense \\
\grayrow
$r_{k,\text{nav}}$ & 
$2.5 \cdot (||\Delta\mathbf{p}_{k-1, \mathbf{G}_\text{opt}}||_2 - ||\Delta\mathbf{p}_{k, \mathbf{G}_\text{opt}}||_2)$ & Dense \\
$r_{k,\text{move}}$ & 
$\begin{cases}
-0.5 & \text{if }\ \max_{j=1,2,3}\|\Delta\mathbf{p}_k-\Delta\mathbf{p}_{k-j}\|_2 < \frac{s_{max}}{4}\\
0 & \text{else}
\end{cases}$ & Dense \\
\grayrow
$r_{k,\text{exp}}$ & 
$-0.05$ & Dense \\
\bottomrule
\end{tabular}
\label{tab:rewards}
\end{table}

In Table \ref{tab:rewards}, the term
\begin{equation*}
\Delta\tilde{\mathbf{p}}_k=\frac{\Delta\mathbf{p}_k}{||\Delta\mathbf{p}_k||_2}
\end{equation*}
denotes the normalized position vector from the agent to the target at time step $k$.
$\mathbf{n}_k$ represents the agent's heading vector at time step $k$.
Both vectors are expressed in the world frame.
The term
$||\Delta\mathbf{p}_{k, \mathbf{G}_\text{opt}}||_2 = ||\Delta\mathbf{p}_{k} - \mathbf{G}_\text{opt}||_2$
corresponds to the Euclidean distance between the agent and the optimal inspection goal point $\mathbf{G}_\text{opt} \in \mathbb{R}^2$. $\mathbf{G}_\text{opt}$ is computed during the dataset generation using the ground truth algorithm introduced in Section~\ref{subsec:ground_truth}.

\begin{figure}[h]
    \centering
    % \includesvg[width=0.95\linewidth]{figures/reward.svg}
    \includegraphics[width=0.9\linewidth]{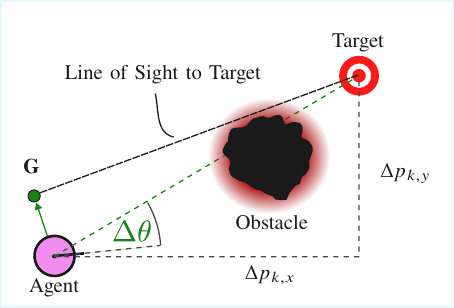}
    \caption{In every step, the agent gets a reward based on the relative orientation to the target and the distance to the ground truth inspection point $\mathbf{G}_{opt}$. These rewards are indicated in green.}
    \label{fig:reward}
\end{figure}

We provide a positive terminal reward $r_{k,T}$ upon successful completion of an episode, where success is defined as reaching a position from which the agent can observe the target without collisions and within the time-limit constraints of $100$ steps.
Conversely, a penalty is imposed and the episode terminates in the event of a collision.
If the episode terminates due to a timeout, the agent receives the same penalty.
These penalty terms encourage the agent to learn a policy that is both safe and efficient.
The agent receives a reward $r_{k, \text{orient}}$ based on the relative orientation $\Delta \theta_k$ to the target, as illustrated in Figure~\ref{fig:reward}.
This reward is computed using the dot product of the agent's heading direction $\mathbf{n}_k$ and the normalized position vector from the agent to the target $\Delta \mathbf{p}_k$, yielding a continuous measure of alignment.
An important advantage of this formulation is that the reward varies smoothly with the agent’s orientation, providing continuous feedback rather than discrete penalties.
In case the reward is positive, it is scaled down to prevent the agent from overfitting to this particular reward term and thus hinder its ability to navigate the environment without colliding.

The reward term $r_{k, \text{nav}}$ is defined as a dense reward signal that encourages the agent to move towards the optimal inspection goal point $\mathbf{G}_\text{opt}$.
It is computed by comparing the Euclidean distance from the previous position $\mathbf{p}_{k-1}$ to $\mathbf{G}_\text{opt}$ with the Euclidean distance from the current position $\mathbf{p}_k$ to $\mathbf{G}_\text{opt}$.
Hence, if the agent chooses an action that reduces the Euclidean distance to the optimal inspection goal point $\mathbf{G}_\text{opt}$, it receives a positive reward.
Conversely, if the agent's action increases the Euclidean distance to $\mathbf{G}_\text{opt}$, the agent receives a penalty.
A key advantage of this formulation is its continuous nature, which provides smooth feedback at every time step $k$ and thus facilitates stable learning.

During policy training, we observed that the agent often exhibits oscillatory behavior, repeatedly rotating in place until a timeout occurs.
This behavior arises because the agent exploits the positive reward provided by $r_{k, \text{orient}}$ when it remains stationary while aligning with the target.
To mitigate this issue, we introduce an additional penalty term, $r_{k, \text{move}}$, which penalizes the agent for remaining within a small neighborhood of its current position for more than three consecutive time steps.
This term encourages the agent to make forward progress rather than indefinitely rotating in place, thereby promoting more efficient action choices.
Moreover, we find that $r_{k, \text{move}}$ is particularly effective in helping the agent escape local minima, as it discourages stagnation at a single position.

Lastly, a small step penalty $r_{k, \text{exp}}$ is added to the total reward $r_k$ to further incentivize exploration and encourage the agent to take a minimal number of actions.
Thus, the total reward $r_k$ provided at each time step $k$ is described by the following function
\begin{align}
    r_k=
    \begin{cases}
        r_{k, T} & \text{if terminate at $k$}\\
        r_{k, \text{orient}} + r_{k, \text{nav}} + r_{k, \text{move}}+r_{k, \text{exp}} & \text{else}
    \end{cases}
\end{align}
\subsection{Model Architecture}
\label{subsec:architecture}
Figure~\ref{fig:architecture} provides an overview of the implemented policy architecture for our reinforcement learning pipeline.  
At every time step $k$, the agent receives an ego-centric depth image $\mathbf{I}_k$, the position vector from the agent to the target $\Delta \mathbf{p}_k$, the relative orientation of the agent to the target $\Delta \theta_k$ and the past three actions $\mathbf{u}_{k-1}, \mathbf{u}_{k-2}, \mathbf{u}_{k-3}$ as an input.
A custom feature extractor is used to encode spatial and directional information from the agent’s observations into a compact latent representation suitable for policy learning.
A three-layer Convolutional Neural Network (CNN)~\cite{cnn} maps the depth image into a latent feature space, while the remaining inputs are transformed using Single-Layer Perceptrons (SLPs) into their respective latent representations.
These feature vectors are then concatenated to form a unified latent state, which is provided to the policy network.
Proximal Policy Optimization (PPO)~\cite{ppo} is employed to train the agent with a continuous action space due to its stability, sample efficiency, and ability to handle continuous action and observation spaces.
The action $\mathbf{u}_k$ is subsequently passed to a low-level controller, which executes the corresponding movement in the environment.
\section{Experiment and Result}
\label{sec:results}
\begin{table*}[ht]
    \centering
\caption{Qualitative results under: (i) collisions and wall-sliding enabled (C\&S), (ii) collisions enabled, wall-sliding disabled (C\&NS), and (iii) both disabled (NC\&NS).}
    \label{results_coll_wall_comb}
    \begin{center}
        \begin{tabular}{ccccccc}
            \toprule
            \textbf{Method} & \textbf{SR (C\&S)} & \textbf{SPL (C\&S)} & \textbf{SR (C\&NS)} & \textbf{SPL (C\&NS)} & \textbf{SR (NC\&NS)} & \textbf{SPL (NC\&NS)}\\
            \midrule
            \grayrow
            Random & 20.67\%  & 0.0628 & 15.87\% & 0.0302 & 5.53\% & 0.0264\\
            Obstacle Avoider & 13.7\% & 0.1017 & 13.7\% & 0.1017 & 13.7\% & 0.1017\\
            \grayrow
            DD-PPO & \textbf{93.75\%} & 0.6224 & 74.28\% & 0.3351 & 19.61\% & 0.1233\\
            \midrule
            Ours & 92.07\% & \textbf{0.6657} & \textbf{90.38\%} & \textbf{0.6536} & \textbf{81.49\%} & \textbf{0.6021}\\
            \bottomrule
        \end{tabular}
    \end{center}
\end{table*}
\subsection{Experimental Setup}
\label{subsec:dataset}
We use the Habitat simulator~\cite{habitat19iccv} with the Gibson dataset~\cite{gibsonenv} to train and evaluate our policy in realistic indoor environments. We focus on short-range inspection scenarios with start-to-target distances sampled uniformly between 3.5 and 4.5 meters. This range reflects typical indoor layouts and allows meaningful comparison between inspection and navigation behaviors. Specifically, the start ($\mathbf{S}_i$) and target ($\mathbf{T}_i$) positions for each episode are sampled uniformly from the traversable space $D_{\text{free}} \subset D$, constrained by:
\[
3.5\text{m} \leq ||\mathbf{S}_{i,\text{proj}} - \mathbf{T}_{i,\text{proj}}||_2 \leq 4.5\text{m},
\]
where $||\cdot||_2$ denotes Euclidean distance on the 2D floor plane.

To ensure diverse yet non-trivial episodes, we apply rejection sampling to keep the geodesic-to-Euclidean distance ratio within $[1.1, 1.5]$, following standard metrics used in navigation~\cite{habitat19iccv}. We analyze the shortest possible inspection path in each episode and find that agents travel, on average, 2.1 meters before gaining visibility of the target. In contrast, following the shortest A*-based navigation path until the target becomes visible results in an average of 2.45 meters. This supports our hypothesis that explicitly optimizing for visibility leads to more efficient inspection trajectories.

Training was conducted on an NVIDIA GeForce RTX 4090 for a total of 10 million time steps using 10 parallel environments, resulting in a wall-clock time of about two hours.

We test our policy on 400+ episodes across 14 previously unseen environments and compare it against three baselines: 
(1) a random policy, 
(2) an obstacle-avoidance policy inspired by the Goal Follower baseline~\cite{habitat19iccv}, and 
(3) DD-PPO~\cite{Wijmans2020DD-PPO}, a state-of-the-art RL-based point-goal navigation policy. We picked a provided pre-trained version, that is trained on the Gibson-4+ subset, which is identical to our training set. We evaluate performance using two common metrics for navigation~\cite{on-evaluation-of-embodied-navigation-agents}:
\begin{itemize}
    \item \textbf{Success Rate (SR)}: The percentage of episodes where the agent successfully observes the target, defined as:
    \begin{equation}
        \text{SR} = \frac{1}{N} \sum_{i=1}^N s_i,
        \label{sr}
    \end{equation}
where $s_i = 1$ if the target is visible in episode $i$, and $0$ otherwise.

    \item \textbf{Success-weighted Path Length (SPL)}:
    \begin{equation}
        \text{SPL} = \frac{1}{N} \sum_{i=1}^N s_i \; \frac{l_i}{p_i},
        \label{spl}
    \end{equation}
where $l_i$ is the shortest inspection path and $p_i$ is the agent’s actual path length.
\end{itemize}

To ensure fairness, all episodes terminate once the agent gains clear visual access to the target. Visibility is determined by projecting the 3D target location onto the camera plane and comparing its depth value to the agent’s range, following the observability condition introduced in Section~\ref{subsec:ground_truth}.

\subsection{Simulation Results}
\label{subsec:sim_results}
Table~\ref{results_coll_wall_comb} presents the quantitative results across three evaluation settings: (i) collisions and wall-sliding enabled, (ii) collisions enabled but wall-sliding disabled, and (iii) neither collisions nor sliding allowed. The latter two reflect more realistic deployment scenarios, with the third setting being the most desirable. The first setting is specific to the simulator but is included to allow fair comparison with DD-PPO, which was trained under those conditions.

In the setting with both collisions and wall-sliding enabled, DD-PPO achieves a higher success rate (SR) than our policy. However, our method yields a higher Success-weighted Path Length (SPL), indicating that it produces more efficient inspection paths. In the second evaluation setting, where collisions are allowed but wall-sliding is disabled, DD-PPO's performance declines relative to the first scenario. This suggests that DD-PPO leverages simulator-specific sliding dynamics. In contrast, our policy demonstrates greater robustness, maintaining similar SR and SPL values to those observed in the first setting. Finally, in the third setting, where collisions are entirely prohibited, our policy continues to perform well, achieving a success rate of 81.49\% and demonstrating strong collision avoidance capabilities. It is also worth noting that, unlike in point-goal Navigation tasks where the agent must explicitly issue a \textit{Terminate} action to indicate task completion, we define success based on whether the target becomes visible to the agent. This criterion can inflate the success rate of random policies, as success does not require deliberate stopping behavior.
\begin{figure}[t]
    \centering
    \includegraphics[width=0.98\linewidth]{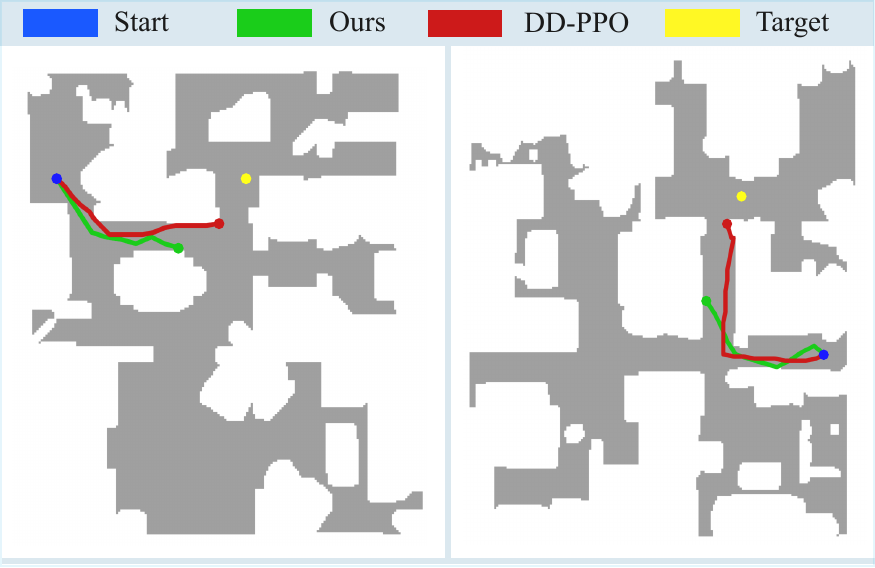}
    \caption{
    Qualitative comparison between our policy and DD-PPO on test episodes. Traversable areas are shown in gray, while obstacles are marked in white. Our policy prioritizes maintaining visibility of the target, while DD-PPO focuses on minimizing navigation distance, often at the expense of target observability.
    }
    \label{fig:qualitative_results}
\end{figure}

Figure~\ref{fig:qualitative_results} presents two qualitative examples. Our policy demonstrates clear advantages by taking shorter and more efficient paths to inspect the target. Unlike DD-PPO, which tends to follow the shortest navigation path that often stays close to the boundaries of the free space and results in occluded views of the target, our policy avoids such behavior. As a result, DD-PPO frequently requires a longer trajectory before the target becomes visible. In contrast, our policy effectively balances navigation and visibility. It not only moves toward the vicinity of the target but also actively seeks locally nearby viewpoints that are more likely to offer a clear line of sight. Additionally, our policy exhibits strong collision avoidance. Unlike DD-PPO, which often collides with and slides along walls, our agent navigates cleanly through cluttered environments.

\subsection{Real-World Experiments}
\label{subsec:real_results}
To evaluate the sim-to-real transferability of our policy, we deploy it on a Boston Dynamics Spot quadruped equipped with an NVIDIA Jetson Orin.
For baseline comparison, we deploy the DD-PPO policy on the same platform.
At the start of each test episode, a reference target point is specified.
To provide the policy with the position vector from the agent to the target $\Delta \mathbf{p}_k$ and relative orientation $\Delta \theta_k$ at each inference step $k$, we leverage Spot’s onboard odometry, which tracks its position relative to the episode’s starting point.
Depth images are captured using a ZED 2i camera and provided as input to the policy.
Unlike the noise-free measurements available in the Habitat simulator, real-world measurements are inherently noisy.
To mitigate its effect when checking the observability condition, we average the depth values within a $5 \times 5$ window centered on the projected goal point.

We conduct two experiments in which the robot is tasked with inspecting a designated target point.
In both scenarios, the direct line of sight to the target is initially occluded:
    \begin{enumerate}
    \item Comparison of DD-PPO and our policy to highlight differences in inspection behavior.
    \item Deployment of our policy under different initial conditions to evaluate its robustness.
\end{enumerate}
In both cases, our policy demonstrates the intended \emph{inspection behavior}, characterized by the robot minimizing unnecessary movement.
A distinct difference between the behavior of DD-PPO and our policy becomes evident, as DD-PPO executes the shortest navigation path and our policy minimizes movement by finding a shorter path to a position from which the target can be observed.
Notably, we deploy our policy zero-shot without retraining.

\begin{figure}[t]
    \centering
    \includegraphics[width=\linewidth]{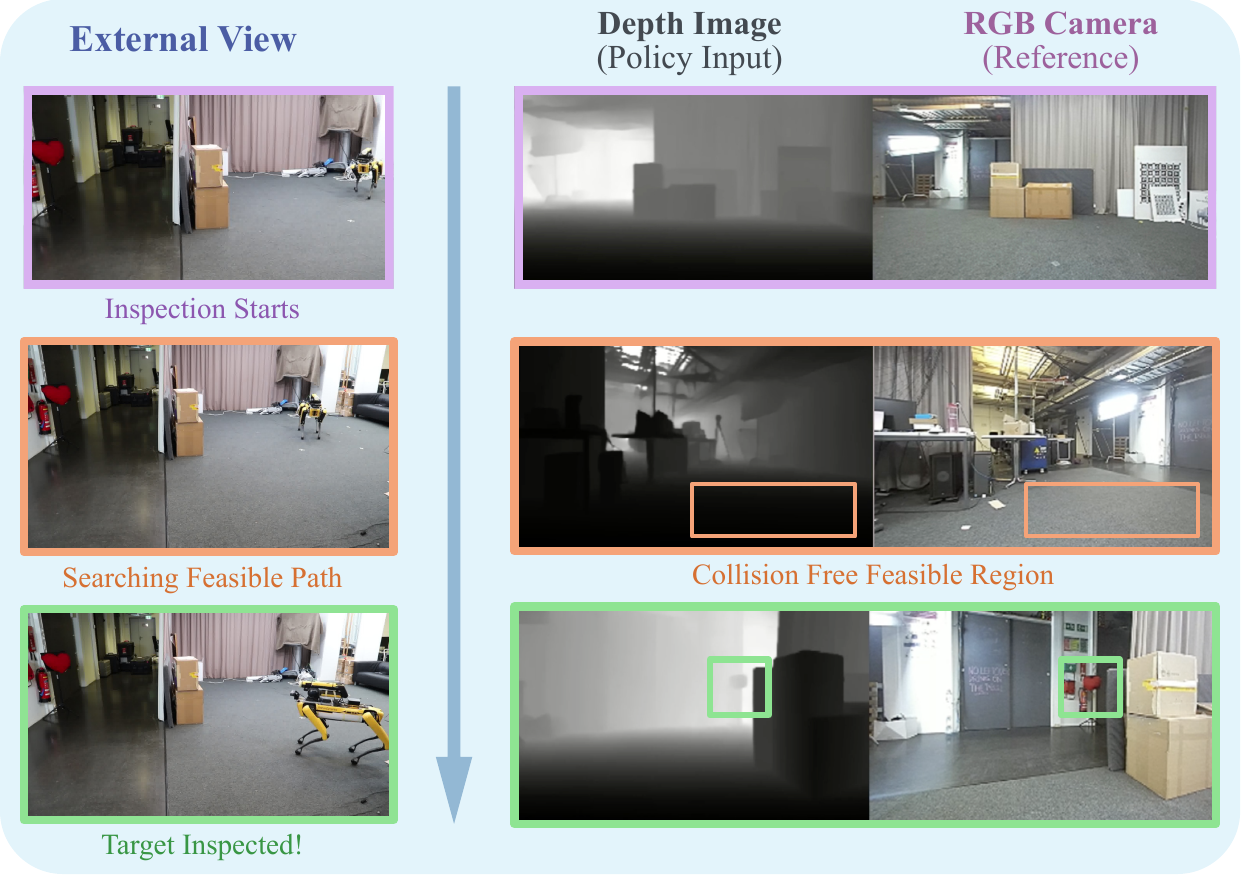}
    \vspace{-0.3cm}
    \caption{
    Real-world experiments of our proposed system were conducted on a physical robot.
    On the left, we present external views showing how the robot gradually navigates to perceive the target without relying on explicit mapping.
    On the right, we visualize the onboard sensing configuration for both the depth and RGB cameras.
    }
    \label{fig:real_world}
\end{figure}

\section{Conclusion}
\label{sec:conclusion}
In this work, we focus on autonomous inspection, which requires a robot to obtain visual access to a target as efficiently as possible. Although this task is critical for many real-world applications, it is often treated as a navigation problem, leading to sub-optimal solutions. We begin by formalizing the inspection task and introducing a theoretical metric to evaluate inspection optimality. We then propose a reinforcement learning-based policy that explicitly incorporates visibility as a training objective. Our approach does not rely on any global map and is trained entirely in simulation. Despite this, it outperforms a state-of-the-art RL-based navigation policy in target inspection tasks. Finally, we demonstrate that our trained policy generalizes zero-shot to a real-world legged robot, showing strong performance in practical inspection scenarios.

%
% \section*{ACKNOWLEDGMENT}
%
% \balance
\bibliographystyle{IEEEtran}
\bibliography{icra}
% \clearpage
%
% \input{chapters/appendix}
\end{document}